\documentclass[letterpaper]{article} 
\usepackage{aaai25}  
\usepackage{times}  
\usepackage{helvet}  
\usepackage{courier}  
\usepackage[hyphens]{url}  
\usepackage{graphicx} 
\usepackage{natbib}  
\usepackage{caption} 
\usepackage[utf8]{inputenc} 
\usepackage[T1]{fontenc}    
\usepackage{booktabs}       
\usepackage{amsfonts}       
\usepackage{nicefrac}       
\usepackage{microtype}      
\usepackage{xcolor}         
\usepackage{subcaption}
\usepackage{amsmath,amssymb,amsfonts}
\usepackage{braket}
\usepackage{mathtools}
 \usepackage{multirow}
 \usepackage{multicol}
\usepackage{algorithm}
\usepackage{algorithmic}
\usepackage{newfloat}
\usepackage{listings}
\DeclareCaptionStyle{ruled}{labelfont=normalfont,labelsep=colon,strut=off} 
\floatstyle{ruled}
\newfloat{listing}{tb}{lst}{}
\floatname{listing}{Listing}
\title{A Scalable Quantum Non-local Neural Network for Image Classification}
\author{
    Sparsh Gupta\textsuperscript{\rm 1},
    Debanjan Konar\textsuperscript{\rm 2},
    Vaneet Aggarwal\textsuperscript{\rm 2}
}
\affiliations{
    \textsuperscript{\rm 1}Franklin W. Olin College of Engineering, Needham, MA 02492 USA\\
    \textsuperscript{\rm 2}Purdue University, West Lafayette, IN 47907 USA\\
    sgupta1@olin.edu, dkonar@purdue.edu, vaneet@purdue.edu
}
\begin{document}
\maketitle
\begin{abstract}
Non-local operations play a crucial role in computer vision, enabling the capture of long-range dependencies through weighted sums of features across the input, surpassing the constraints of traditional convolution operations that focus solely on local neighborhoods. Non-local operations typically require computing pairwise relationships between all elements in a feature set, leading to quadratic complexity in terms of time and memory. Due to the high computational and memory demands, scaling non-local neural networks to large-scale problems can be challenging. This article introduces a hybrid quantum-classical scalable non-local neural network, the Quantum Non-Local Neural Network (QNL-Net), to enhance pattern recognition. The proposed QNL-Net relies on inherent quantum parallelism to allow the simultaneous processing of a large number of input features, enabling more efficient computations in quantum-enhanced feature space and involving pairwise relationships through quantum entanglement. We benchmark our proposed QNL-Net with other quantum counterparts for binary classification with the MNIST and CIFAR-10 data sets. The simulation findings showcase our QNL-Net, which achieves cutting-edge accuracy levels in binary image classification among quantum classifiers while utilizing fewer qubits.
\end{abstract}
%
\begin{links}
     \link{Code}{https://github.com/sparshgup/QNL-Net}
\end{links}

\section{Introduction}
\label{sec:intro}

Computer vision, a fundamental component of Artificial Intelligence (AI), encompasses various applications such as autonomous driving \cite{bao2023heat}, medical imaging~\cite{konar2023_tnnls}, and healthcare~\cite{zhou2022computer}. Image classification is a crucial task within computer vision, with the aim of assigning labels to images based on their visual content. This task serves as the foundation for more complex applications like image segmentation, object detection, and scene understanding. Convolutional Neural Networks (CNNs) have significantly advanced image classification, achieving top-tier performance on datasets like ImageNet~\cite{alexnet2012}. However, CNNs are restricted by their local receptive fields, limiting their ability to capture extensive contextual information and long-range dependencies in images. To address these limitations, non-local neural networks were introduced to capture long-range dependencies, building upon the self-attention mechanism seen in Transformer architectures~\cite{attention_NIPS2017}. Non-local neural networks have proven effective in capturing global context and improving performance in computer vision tasks that require modeling long-range dependencies~\cite{wangnonlocal}. These networks have shown significant improvements in tasks such as image classification, where understanding broader contextual information is crucial for accurate labeling~\cite{shi2021}. Moreover, non-local operations typically require computing pairwise relationships between all elements in a feature set, leading to quadratic complexity in terms of time and memory. Due to the high computational and memory demands, scaling non-local neural networks to large-scale problems is a challenging proposition.

Quantum Machine Learning (QML) has emerged as a revolutionary technology that combines principles of quantum mechanics with Machine Learning (ML), offering transformative potential in various fields~\cite{Biamonte_NatureQML2017}. Unlike classical computing that relies on bits representing $\ket{0}$ and $\ket{1}$, Quantum Computing (QC) introduces qubits that can exist in quantum states simultaneously, leveraging superposition and quantum parallelism for enhanced computational capabilities~\cite{nielsen2010quantumbook}. This quantum advantage allows the development of algorithms that can efficiently tackle complex challenges, potentially revolutionizing areas such as pattern recognition, image classification, optimization, and cryptography~\cite{pqcnature2017}. Within QML, algorithms such as Quantum Support Vector Machine (QSVM), Quantum Kernel methods, and Variational Quantum Classifiers (VQC) have been introduced for classification tasks~\cite{Benedetti_2019}. VQC, in particular, stands out as it combines classical and quantum computing utilizing a quantum circuit with parameterized quantum gates. These parameters can be optimized using classical methods, enabling a hybrid quantum-classical approach to effectively train models~\cite{ding2024}.

Researchers have actively investigated the potential advantages of QML over classical machine learning, both theoretically and practically~\cite{Biamonte_NatureQML2017}. A specific sub-domain within this field, Quantum Neural Networks (QNN), has been extensively developed to evaluate their ability to surpass classical neural networks. Recent studies have used simulations on actual quantum hardware to demonstrate the benefits offered by QML~\cite{naturecompqnnabbas2021}. Despite these advances, challenges persist in QML, including limitations in fault-tolerance, evolving quantum error correction techniques, and issues related to quantum scalability~\cite{nisqpreskill2018,gujju2023quantum}. These challenges present opportunities for innovative research in transitioning from the current Noisy Intermediate-Scale Quantum (NISQ) era to a new era of quantum computing~\cite{huang2021}. The limitations in fault-tolerance and quantum scalability impede the practical applications of QML algorithms. However, ongoing research efforts to develop new quantum algorithms and address these challenges provide a platform to test methodologies on a smaller scale and pave the way for future applications as quantum processors become more scalable. 

In this study, we introduce a novel Quantum Non-local Neural Network (QNL-Net) that merges QC principles with the non-local neural network mechanism to tackle pairwise non-local operations in a feature set harnessing the inherent characteristics of quantum entanglement. The proposed QNL-Net establishes non-local correlations through quantum-enhanced features, emulating classical non-local operations while leveraging quantum entanglement to enhance model performance and capabilities. The proposed QNL-Net architecture introduces a novel approach that efficiently utilizes fewer qubits, a crucial aspect in the NISQ era. Beyond this efficiency, the architecture's significance lies in its treatment of rotations around axes and entanglement. The selection of rotation gates in the proposed QNL-Net ansatzes is closely linked to the core concept of non-local neural networks, which aim to capture intricate spatial dependencies within data. In the quantum realm, rotation gates play a pivotal role in achieving this objective by maneuvering quantum states around various axes, thereby facilitating spatial transformations.

The primary objective of this research is to improve the recognition of patterns and binary classification tasks in computer vision by effectively capturing long-range dependencies. QML holds the potential to overcome several drawbacks of classical non-local neural nets, leveraging the unique characteristics of QC. Quantum parallelism allows for the simultaneous processing of a large number of states, enabling more efficient computations in tasks involving pairwise relationships. Moreover, quantum feature mapping in the proposed QNL-Net transforms classical data into high-dimensional quantum spaces where patterns become more interpretable. This helps in understanding the interactions captured by non-local operations. The main contributions of our work are as follows: 
\begin{enumerate}
    \item We introduce a scalable Quantum Non-local Neural Network (QNL-Net) to tackle pairwise non-local operations in a feature set harnessing the inherent characteristics of quantum entanglement.
    \item We also employed classical machine learning techniques for dimensionality reduction of features before these are processed by our proposed QNL-Net framework, therefore leveraging the advantages of hybrid quantum-classical machine learning models.
    \item Our hybrid classical-quantum models, especially CNN-QNL-Net, achieved high nineties classification accuracy on the MNIST dataset (99.96\% test accuracy) and outperformed the benchmark quantum classifiers QTN-VQC ($98.6\%$ with $12$ qubits)~\cite{qtnvqc2023} and hybrid TTN-MERA ($99. 87\%$ with $8$ qubits)~\cite{Grant2018-dy} using significantly fewer qubits ($4$ qubits).
\end{enumerate}


\section{Related Work}
\label{sec:liter_stud}

Image classification using Quantum Machine Learning (QML) has attracted significant attention due to its potential advantages over classical methods. One notable approach is Quantum Convolutional Neural Networks (QCNN), which utilize quantum circuits for convolutional operations, emphasizing quantum phase recognition and error correction optimization techniques \cite{Cong2019}. Another significant development is the Quanvolutional Neural Network, which introduces quantum convolution layers to transform classical data using random quantum circuits for feature extraction, which showcases superior accuracy and training performance compared to classical CNNs \cite{henderson2019quanvolutional}. QML models have shown effectiveness in binary classification tasks for noisy datasets and images~\cite{Schetakis2022}.
Nevertheless, recent advances include notable work that loads matrices as quantum states and introduces trainable quantum orthogonal layers adaptable to various quantum computer capabilities, showing promising results on superconducting quantum computers~\cite{Cherrat2024quantumvision}. These developments highlight the potential of QML to improve image classification systems. 

Various studies have explored the application of Quantum Neural Networks (QNN) in binary classification tasks within computer vision, providing benchmarks for research outcomes. For example, QTN-VQC~\cite{qtnvqc2023} integrated quantum circuits for tensor-train networks with variational quantum circuits to establish an efficient training pipeline. Furthermore, hierarchical quantum classifiers use expressive circuits to classify highly entangled quantum states and demonstrate resilience to noise~\cite{Grant2018-dy}. \citet{konar2023_sqnn} proposes a scalable approach for Spiking Quantum Neural Networks (SQNN) for classification, suggesting the utilization of multiple small-scale quantum devices to extract local features and make predictions. \cite{Jiang2021-of} presented a Quantum Flow model that represented data as unitary matrices to achieve a quantum advantage, reducing the cost complexity of unitary matrices-based neural computation. These recent advances in QML models have shown robustness in image classification, which is proven effective in handling higher-dimensional data more efficiently than its classical counterparts.

\section{Non-local Neural Networks}
\label{sec:NL-Net}

Classical convolution operations in Convolutional Neural Networks (CNN) are widely used in computer vision models but face challenges related to processing long-range dependencies due to the incremental growth in the receptive field with repeated operations. This leads to computational inefficiency and optimization difficulties. To address these limitations, classical Non-local Neural Networks (NL-Net) have been introduced, which incorporate non-local operations that compute responses by aggregating features from all positions in the input, enabling efficient capture of long-range dependencies crucial in computer vision tasks~\cite{wangnonlocal}. 

A generic non-local operation for an input signal's (image, sequence, video) feature map $\mathrm{x} \in \mathbb{R}^{N \times C}$, where $N$ is the number of positions (i.e., pixels) and $C$ is the number of channels, is defined as:
\begin{eqnarray} \label{eqn:1}
    \mathrm{y}_i = \frac{1}{\mathcal{C}(\mathrm{x})} \sum_{\forall j} f(\mathrm{x}_i, \mathrm{x}_j) g(\mathrm{x}_j),
\end{eqnarray}
where $i$ is the index of an output position (in space/time/spacetime) whose response is to be computed, $j$ enumerates over all possible positions, and $\mathrm{y}$ is the output signal. $f$ is a pairwise function that computes a scalar representing the relationship (such as similarity) between $i$ and all $j$. $g$ is a unary function that computes a representation of the input signal at $j$ and normalizes it by a factor $\mathcal{C} (\mathrm{x})$.

In terms of the functions, $g$ is simply considered as a linear embedding such that $g(\mathrm{x}_i) = W_g \mathrm{x}_j$, where $W_g$ is a learned weight matrix. There are several choices for the pairwise function $f$ such as: 
\begin{enumerate}
    \item Gaussian: 
    \begin{eqnarray*}
        f(\mathrm{x}_i, \mathrm{x}_j) = e^{\mathrm{x}_i^T \mathrm{x}_j}, \hspace{1mm} \mathcal{C}(\mathrm{x}) = \sum_{\forall j} f(\mathrm{x}_i, \mathrm{x}_j);
    \end{eqnarray*}
    \item Embedded Gaussian: 
    \begin{eqnarray*}
        f(\mathrm{x}_i, \mathrm{x}_j) = e^{\theta(\mathrm{x}_i)^T \phi(\mathrm{x}_j)}, \hspace{1mm}\mathcal{C}(\mathrm{x}) = \sum_{\forall j} f(\mathrm{x}_i, \mathrm{x}_j);
    \end{eqnarray*}
    where $\theta(\mathrm{x}_i) = W_\theta \mathrm{x}_i$ and $\phi(\mathrm{x}_j) = W_\phi \mathrm{x}_j$ are linear embeddings. This also relates to self-attention~\cite{attention_NIPS2017}, in which NL-Net is just an extension of the computer vision domain, or more specifically, a generic space or spacetime domain. This is due to the fact that for any $i$, $\frac{1}{\mathcal{C}(\mathrm{x})} f(\mathrm{x}_i, \mathrm{x}_j)$ becomes the softmax computation along the dimension $j$. 
    \item Dot Product:
    \begin{eqnarray*}
        f(\mathrm{x}_i, \mathrm{x}_j) = \theta(\mathrm{x}_i)^T \phi(\mathrm{x}_j), \hspace{1mm} \mathcal{C}(\mathrm{x}) = N;
    \end{eqnarray*}
    where $N$ is the number of positions in $\mathrm{x}$.
    \item Concatenation:
    \begin{eqnarray*}
        f(\mathrm{x}_i, \mathrm{x}_j) = ReLU(w_f^T [\theta(\mathrm{x}_i)^T, \phi(\mathrm{x}_j)]), \mathcal{C}(\mathrm{x}) = N;
    \end{eqnarray*}
\end{enumerate}
A non-local block in spacetime encapsulates the non-local operation in Eq. ~\eqref{eqn:1} elegantly and is defined as:
\begin{eqnarray}
    \mathrm{z_i} = W_i \mathrm{y}_i + \mathrm{x}_i
\end{eqnarray}
where $\mathrm{y}_i$ is the non-local operation and residual connection `$+ \hspace{1mm} \mathrm{x}_i$' allows integrating the non-local block into any pre-trained model without disruptions in their initial behavior~\cite{residual_he2015deep}.

\section{Quantum Non-local Neural Network} 
\label{QNL-Net}

In this study, we present the Quantum Non-Local Neural Network (QNL-Net), which leverages trainable quantum circuits to implement non-local operations, effectively capturing and processing long-range dependencies within input data. Our proposed QNL-Net architecture integrates classical dimensionality reduction techniques to operate as a hybrid quantum-classical classifier. This section entails the design and implementation of the QNL-Net architecture, followed by the incorporation of classical dimensionality reduction methods to create hybrid classifiers, specifically highlighting the CNN-QNL-Net and PCA-QNL-Net models. Additionally, we explore post-QNL-Net classical computation to enhance the classification process further.

The proposed QNL-Net introduces a novel approach by translating classical non-local operations into quantum circuits, harnessing the parallelism and entanglement properties inherent in quantum computing. This translation process involves the design of quantum gates and circuits that mimic the functionality of classical non-local layers, enabling the network to analyze complex data structures more efficiently.
\begin{figure}[t]
    \centering
    \includegraphics[width=84mm]{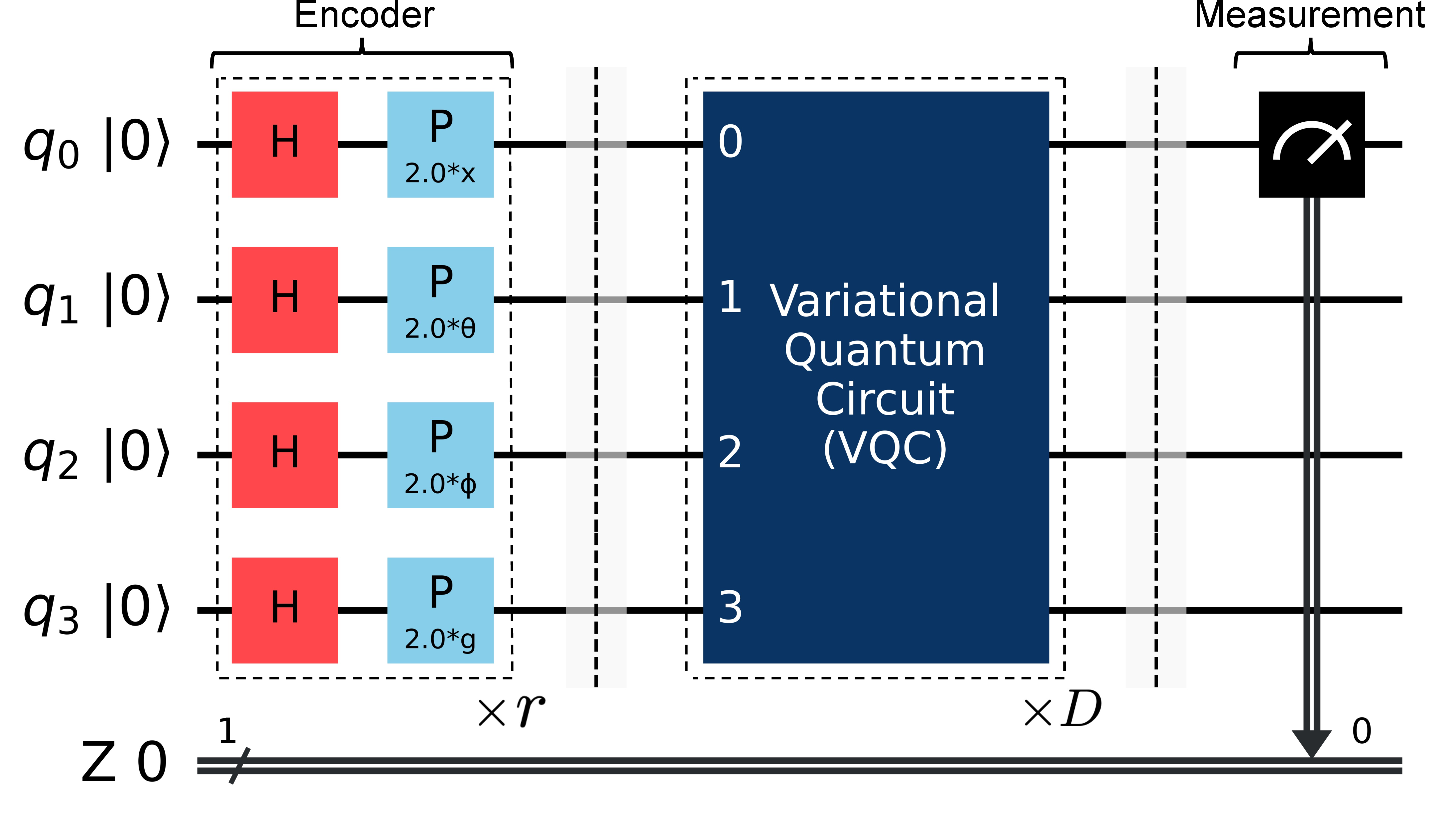}
    \caption{The Quantum Non-local Neural Network (QNL-Net) comprises a four-qubit circuit composed of three parts: (i) Encoder: To encode classical data into quantum states. (ii) Variational Quantum Circuit (VQC): classically trainable quantum circuit. (iii) Measurement: the circuit is measured at qubit $0$ in the Pauli-$Z$ basis. The encoder and the VQC ansatz have $r$ and $D$ repetitions respectively. The Encoder has $4r$ trainable parameters and the VQC has $5D$ trainable parameters (for $n=4$ input qubits).}
    \label{fig:circuit}
\end{figure}
To encode  classical data $X = [y_0, y_1, \cdots, y_{n-1}] \in \mathbb{R}^n$ into the quantum space, we first write $\ket{X}$ as the quantum version of $X$ as $\ket{X} = \ket{y_0, y_1, \cdots, y_{n-1}}$. Then, we encode this data as
\begin{eqnarray}
    \ket{\psi_\Phi} = (  \bigotimes^{n}_{k=1} {P(\lambda_k)} H^{\otimes n})^r \ket{X}, 
\end{eqnarray}
where 
$H^{\otimes n}$ is a layer of Hadamard gates acting on all $n$ qubits and $P(\lambda) = \begin{bmatrix} 1 & 0 \\ 0 & e^{i \lambda} \end{bmatrix}$. Thus, the encoder has $nr$ parameters which can be trained for efficient encoding. 


\begin{figure*}
    \centering
    \includegraphics[width=0.8\textwidth]{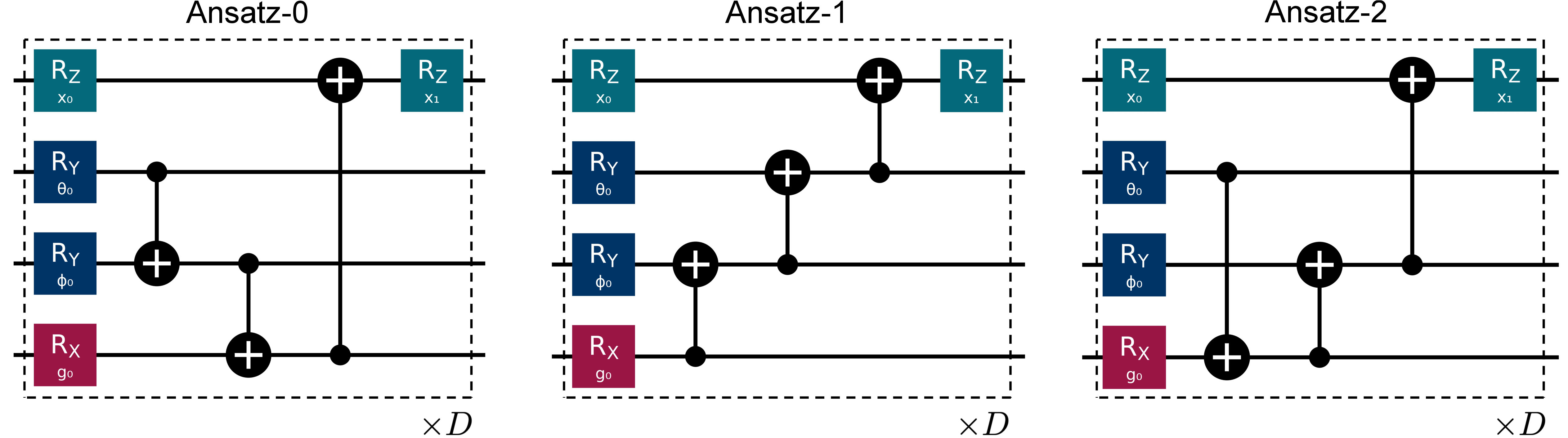}
    \caption{The three ansatzes used as the Variational Quantum Circuits (VQC) in our QNL-Net mechanism. $R_x, R_y, R_z$ rotation gates represent single-qubit rotations along the $x, y, z$ axes, respectively, with trainable parameters. The strategies for performing entanglement using $C_X$ gates (CNOT) are: cyclic pattern (Ansatz-0), reverse linear chain (Ansatz-1), and a mixed pattern (Ansatz-2).}
    \label{fig:ansatz}
\end{figure*}

We employ Variational Quantum Circuits (VQC) in our QNL-Net framework, which are classically trainable to optimize parameters using a predefined cost function~\cite{Benedetti_2019}. This hybrid quantum-classical strategy harnesses the expressive capabilities of quantum circuits while utilizing classical optimization methods for efficient parameter adjustment. Our implementation includes three distinct ansatzes, each incorporating five parameterized rotation gates to provide precise control over the quantum state. These ansatzes feature three $C_X$ gates with unique configurations, establishing entanglement-induced correlations among the qubits to ensure that the quantum state captures non-local dependencies within the data. The quantum circuits for these ansatzes are illustrated in Fig.~\ref{fig:ansatz}. Initially, rotation gates are applied to each qubit in all ansatzes to derive the subsequent quantum state
\begin{equation}
    \ket{\psi_1} = R_z(x_0)_{q_0} R_y(\theta_0)_{q_1} R_y(\phi_0)_{q_2} R_x(g_0)_{q_3} \ket{\psi_\Phi},
\end{equation}
\begin{equation}
\text{such that, }
    R_x(\lambda) = \begin{bmatrix}
        cos(\frac{\lambda}{2}) & \hspace{-2mm}
        \hspace{-2mm} -i\hspace{1mm}sin(\frac{\lambda}{2}) \\
        -i\hspace{1mm}sin(\frac{\lambda}{2}) & \hspace{-2mm}\hspace{-2mm}cos(\frac{\lambda}{2})
    \end{bmatrix},
\end{equation}
\begin{equation}
    R_y(\lambda) = \begin{bmatrix}
        cos(\frac{\lambda}{2}) & \hspace{-2mm} -sin(\frac{\lambda}{2}) \\
        sin(\frac{\lambda}{2}) & \hspace{-2mm} cos(\frac{\lambda}{2})
    \end{bmatrix}, 
    R_z(\lambda) = \begin{bmatrix}
        e^{-i \frac{\lambda}{2}} & \hspace{-3mm} 0 \\
        0 & \hspace{-3mm} e^{i \frac{\lambda}{2}}
    \end{bmatrix}.
\end{equation}

The variation in the ansatzes stems from distinct entanglement strategies outlined below. These configurations ensure comprehensive entanglement among all qubits to capture non-local dependencies across the circuit. In Ansatz-0, the $C_X$ gates form a cyclic pattern, establishing a loop of entanglement among the qubits. Ansatz-1 features $C_X$ gates forming a reverse linear chain, creating a backward sequential entanglement structure. Ansatz-2 introduces a non-linear and unique mixed pattern entanglement strategy utilizing the $C_X$ gates. The resulting quantum states for each respective ansatz are as follows:
\begin{equation}
    \ket{\psi_2^{[0]}} = C_X(q_1 q_2) C_X(q_2 q_3) C_X (q_3 q_0) \ket{\psi_1},
\end{equation}
\begin{equation}
    \ket{\psi_2^{[1]}} = C_X (q_3 q_2) C_X (q_2 q_1) C_X (q_1 q_0) \ket{\psi_1},
\end{equation}
\begin{equation}
    \ket{\psi_2^{[2]}} = C_X (q_1 q_3) C_X (q_3 q_2) C_X (q_2 q_0) \ket{\psi_1},
\end{equation}
\begin{equation}
\text{where }
    C_X = \begin{bmatrix} 
    1 & 0 & 0 & 0 \\ 
    0 & 1 & 0 & 0 \\ 
    0 & 0 & 0 & 1 \\ 
    0 & 0 & 1 & 0 
    \end{bmatrix}.
\end{equation}

Then, we add our final rotation gate on qubit 0 in each ansatz, and obtain the following quantum state.
\begin{equation}
    \ket{\psi_3^{[a]}} = R_z(x_1)_{q_0} \ket{\psi_2^{[a]}},
\end{equation}
where $a$ represents the desired ansatz and $a = 0, 1,$ or $2$.

This sequence of gates and entanglements in the ansatzes constitutes one layer of the respective ansatz and can also be represented by the unitary operator as
\begin{equation}
    U_a(\theta) = \ket{\psi_3^{[a]}}.
\end{equation}

Now, each layer consists of 5 parameters (can be generalized to $n+1$ parameters for $n$ qubits) and can be applied with $D$ repetitions to enhance the expressiveness of the model, resulting in a total of $5D$ trainable parameters from the VQC, and therefore, we obtain the final state of the circuit, 
\begin{equation}
    \ket{\psi_s} = (U_a(\theta))^D \ket{\psi_\Phi},
\end{equation}
where $a$ is the desired ansatz $a$ = $0, 1$, or $2$.

According to the Born rule, measuring any quantum state in the Pauli-$Z$ basis ($\sigma_z$) either collapses into the state $\ket{0}$ with probability $|\alpha|^2$ or into the state $\ket{1}$ with probability $|\beta|^2$ \cite{nielsen2010quantumbook}. In general, the expectation of any observable $\hat{O}$ for a state $\ket{\psi}$ can be denoted as
\begin{eqnarray}
    \braket{\hat{O}} = \bra{\psi}\hat{O}\ket{\psi} = \sum_{i} m_i p_i
\end{eqnarray}
where $m_i$ are the possible measurement values, \emph{i.e.}, the eigenvalues weighted by their respective probabilities $p_i = |\alpha|^2 - |\beta|^2$. Therefore, to measure an observable $O$ on $n$-qubits, we can also represent it as a sum of tensor products of Pauli operators, such that,
\begin{eqnarray}
    O = \sum_{k} \alpha_k P_k, \hspace{3mm} \alpha_k \in \mathbb{R} ,
\end{eqnarray}
where $\alpha_k \in \mathbb{R}$ are real coefficients, and $P_k = P_{k_1} \otimes P_{k_2} \otimes \cdots \otimes P_{k_n}$ are tensor products of Pauli operators $P_{k_i} \in \{I, X, Y, Z\}$ acting on each of the $n$ qubits. In our mechanism, we measure the circuit in the Pauli-$Z$ computational basis, specifically at one qubit, $q_0$, while the rest of the qubits are measured using the Identity ($I$) operation. So, we can denote the measurement of $q_0$ for a quantum state of $\ket{\psi_s}$ as $\braket{Z}$ such that,
\begin{eqnarray}
    \braket{Z} = \bra{\psi_s} U^{\dagger}(\theta) Z_{q0} U(\theta) \ket{\psi_s},
\end{eqnarray}
where $\theta$ represents the parameters of the quantum ansatz, $Z_{q_0}$ represents the Pauli-$Z$ operator acting on qubit $q_0$, $U(\theta)$ is the unitary operator parameterized by $\theta$, and $U^{\dagger}(\theta)$ is the Hermitian adjoint of $U(\theta)$.

\begin{figure*}
 \vspace{-.05in}
    \centering
    \includegraphics[width=.81\textwidth]{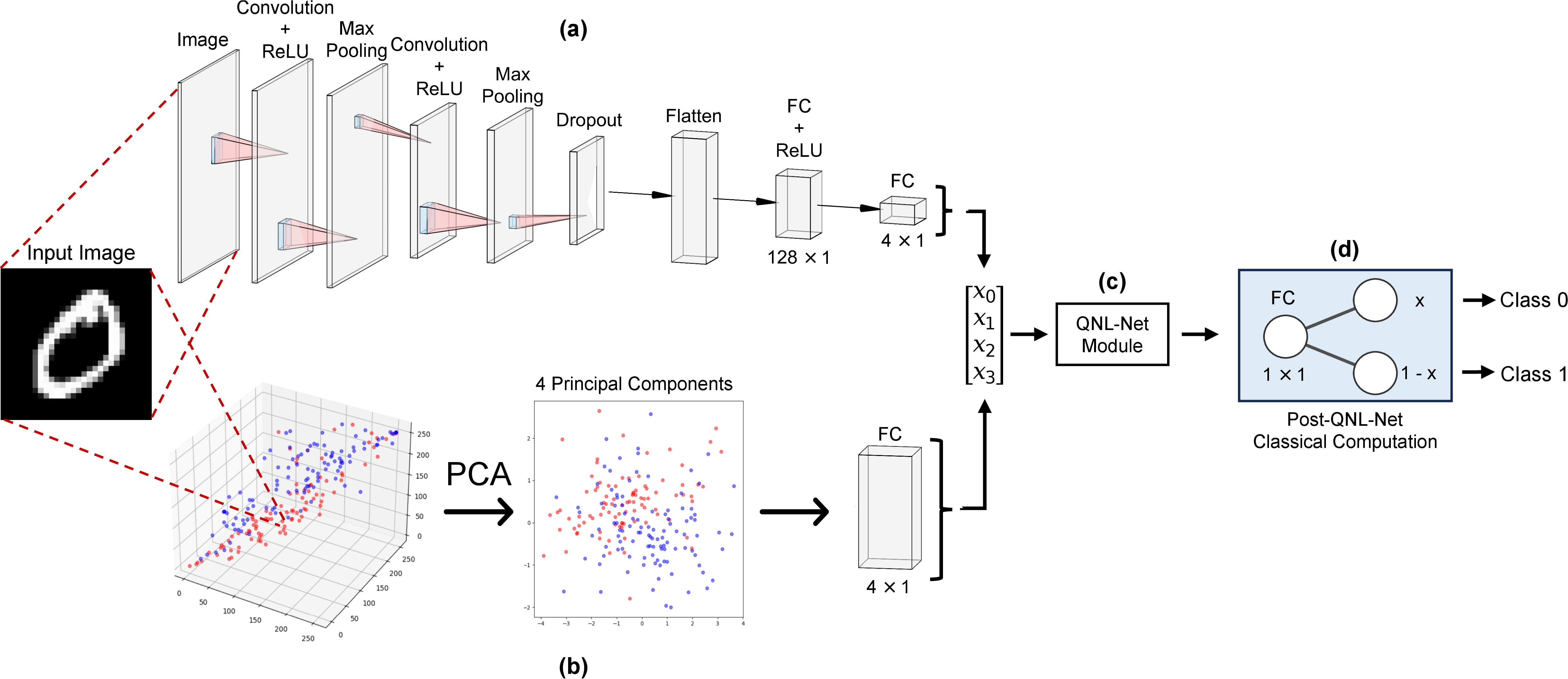}
    \caption{The proposed hybrid Classical-quantum QNL-Net frameworks comprises the following components: (a) CNN-QNL-Net: Two convolutional layers with ReLU activation and max-grouping are used to extract features from input images. Subsequently, a dropout layer is applied, followed by flattening of the features. Two fully connected (FC) layers are utilized to preprocess the data for integration with the QNL-Net. (b) PCA-QNL-Net: Principal Component Analysis (PCA) is utilized to reduce the dimensionality of the input data to 4 components. These components are processed by a fully connected (FC) layer before being input into the proposed QNL-Net. (c) QNL-Net: This component of the pipeline conducts computations on the four features obtained from the classical models within the mechanism and generates a single measurement from qubit 0. (d) Post-QNL-Net Classical Computation: A fully connected layer is employed to fine-tune the quantum output. Subsequently, this output is transformed and concatenated as required for binary classification.}
    \label{fig:model}
\end{figure*}

In this study, we employ two hybrid classical-quantum QNL-Net models: CNN-QNL-Net and PCA-QNL-Net. Classic models are used for dimensionality and feature reduction, converting input data into a feature vector of size $4 \times 1$. Each value in this vector corresponds to one qubit in our QNL-Net layer, facilitating the capture and analysis of long-range dependencies and intricate patterns among these features. The incorporation of CNN architecture into our QNL-Net Module is driven by CNN's ability to capture spatial dependencies and identify local patterns within complex image data through convolutional and pooling layers, effectively reducing input dimensionality while retaining essential features.

In the proposed CNN-QNL-Net architecture, the process begins with two convolutional layers, each integrating an activation function and max pooling, to process an input image tensor $X \in \mathbb{R}^{W \times H \times C}$. Here, $W$ represents the width, $H$ denotes the height, and $C$ signifies the number of channels in the input image (e.g., 1 for grayscale images and 3 for RGB images). The input data matrix is $X \in \mathbb{R}^{N \times P}$, where $N$ is the number of samples in the dataset and $P$ is the total number of pixels per image. The linear transformation performed by this layer is as follows.
\begin{equation}
    x = W_4 Q + b_4,
\end{equation}
where $Q \in \mathbb{R}^1$ is the single QNL-Net output, $W_4 \in \mathbb{R}^{1 \times 1}$ is the weight matrix, and $b_4 \in \mathbb{R}^1$ is the bias vector. 
 Further details on the CNN and PCA architectures are given in Appendix. Our model uses the negative log-likelihood (NLL) loss function for the binary classification problem. The NLL loss measures the variation between the true labels $y$ and the classical predicted probabilities $\hat{y} = [\hat{y}_0, \hat{y}_1]$ obtained from the measurement of the hybrid quantum-classical QNL-Net model, and is defined for binary classification as:
\begin{eqnarray}
    \mathcal{L}(\phi) = 
    -  \sum^{N}_{i=1} \left(y_i \log \hat{y}_{i1} + (1 - y_i) \log \hat{y}_{i0}\right),
\end{eqnarray}
where $N$ is the number of samples in the dataset, $\hat{y}_0$ is the predicted probability for class 0, $\hat{y}_1$ is the predicted probability for class 1, and $\phi$ are the classical parameters. The gradient calculations are described in the Appendix. 

\section{Evaluation Results} 
\label{sec:results}

\subsection{Datasets}
In our Qiskit simulations, we used two widely used image processing datasets for image classification: MNIST~\cite{lecun1998} and CIFAR-10~\cite{alexnet2012} (sample images are shown in Appendix in Fig. \ref{fig:dataset_sample}). For MNIST, we focus on binary classification using the digits $0$ and $1$, comprising $12,665$ training samples and $2,115$ testing samples. For CIFAR-10, we perform a binary classification using classes $2$ (birds) and $8$ (ships). Before feeding the images into the models, we normalize them using the global mean and standard deviation of each data set, scaling the pixel values from [$0, 255$] to [$0, 1$].

\subsection{Simulation Settings}
In our study, we extensively evaluated the performance of the proposed QNL-Net in benchmark datasets, MNIST and CIFAR-10, demonstrating its resistance to noise and potential advantages over traditional quantum binary classification models. The Qiskit simulations are performed on a \emph{MacBook Pro} computer equipped with an \emph{M2 Max} chip and $64$GB RAM. Our QNL-Net has been implemented using the \emph{EstimatorQNN} module of Qiskit Machine Learning $0.7.2$ and Qiskit $1.1.0$ \cite{qiskit2024}, allowing the encoding of classical data into quantum data and facilitating the training of the ansatz. The Qiskit \emph{ZFeatureMap} is utilized to leverage the quantum-enhanced feature space, providing a quantum advantage in classification tasks~\cite{Havlicek2019}. Following the measurement of our QNL-Net output, further classical computation is performed by adding a fully connected layer with a single learnable parameter to fine-tune and optimize the quantum output, thereby enhancing the model's performance. The classical node is constructed and gradient optimization has been performed using \emph{PyTorch} $2.3.0$ \cite{paszke2017automatic}, seamlessly integrated with the \emph{EstimatorQNN} module. The models are trained for $100$ epochs with a batch size of $1$, using the negative log-likelihood (NLL) loss function for convergence. The Adam optimizer \cite{kingma2017adam} is configured with varying learning rates between $0.0001$ and $0.0004$, depending on the model and ansatz used, as detailed in Appendix Table~\ref{table:results}. The \emph{ExponentialLR} scheduler~\cite{li2019exponential} with a decay rate of $0.9$ is used, while the remaining parameters were set to default.

\subsection{Simulation Results}
\begin{figure}[t]
    \centering
    \includegraphics[width=84mm]{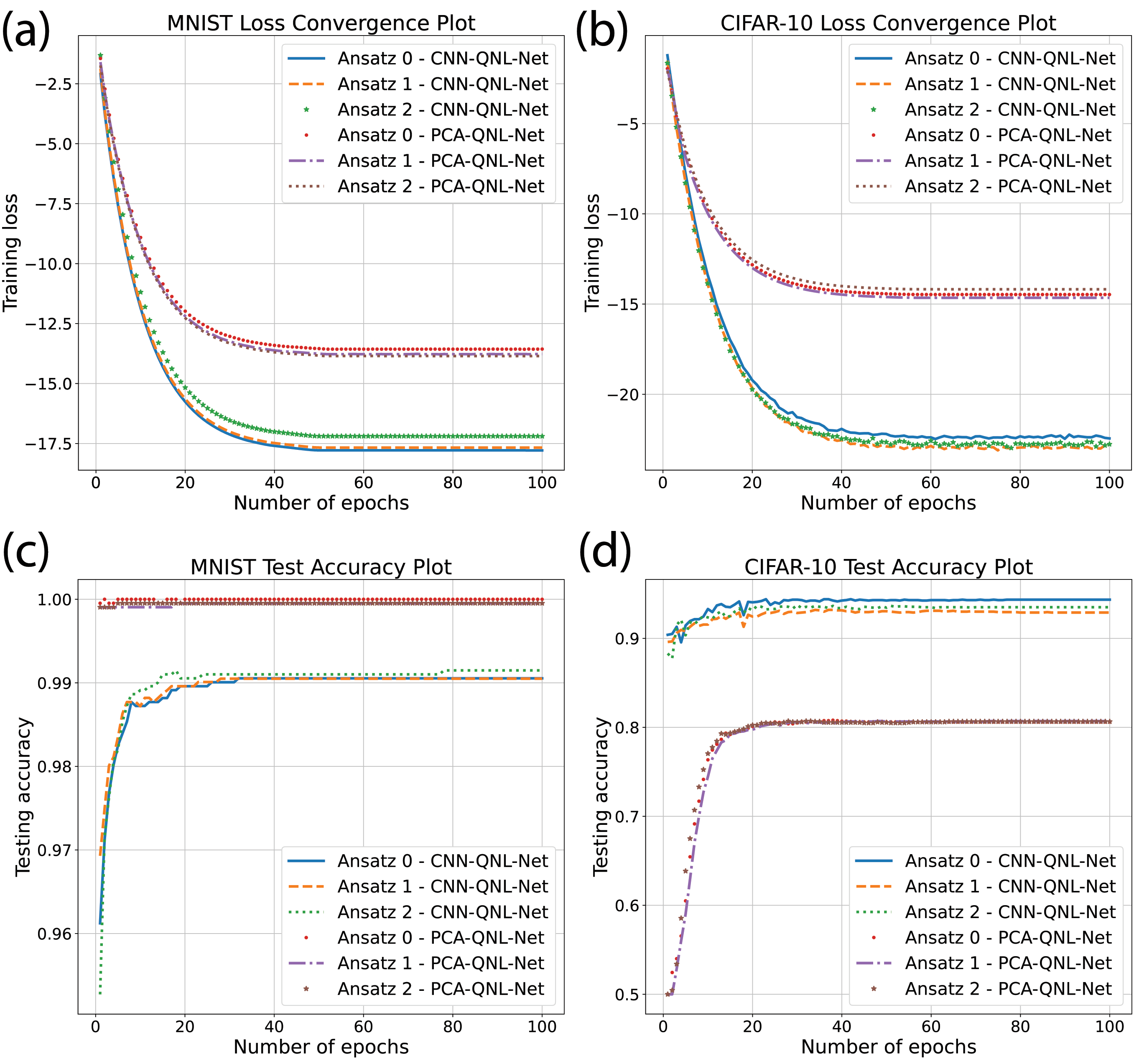}
    \caption{Training loss convergence and test accuracy plots for the CNN-QNL-Net and PCA-QNL-Net models for the three ansatzes with one feature map repetition ($r = 1$) and one ansatz repetition ($D = 1$). (a) and (b) display the training loss convergence on MNIST and CIFAR-10, respectively. (c) and (d) show the corresponding test accuracy on these datasets.}
    \label{fig:results_plots}
\end{figure}

Simulations are performed using all combinations of the number of repetitions for the encoder ($r$ = $1, 2$ or $3$) and the number of repetitions for the VQC ($D$ = $1, 2$ or $3$), with results averaged across the nine possible configurations. The averaged accuracies  across all runs for each specific ansatz and model configuration are given in  Table~\ref{table:results}. The learning rates are found by grid search, and given in Table~\ref{table:results} for these settings. Fig.~\ref{fig:results_plots} illustrate the models' performance and convergence behavior during training for $D=r=1$. The simulation results obtained from the MNIST dataset for classes $0$ and $1$ reveal that our CNN-QNL-Net model marginally outperforms our PCA-QNL-Net model, achieving an average classification test accuracy of $99.96\%$. In contrast, PCA-QNL-Net achieved a test precision of $99. 59\%$. Ansatz-0 and Ansatz-2 generally exhibit superior performance compared to Ansatz-1 for this dataset, as indicated in  Table \ref{table:results}. However, for the CIFAR-10 dataset, hybrid QNL-Net models demonstrate relatively lower performance compared to MNIST due to the inclusion of three color channels (RGB) as opposed to the grayscale images in MNIST. Nonetheless, our CNN-QNL-Net model still achieves an average test accuracy of $93.98\%$. Ansatz-1 performs better on the testing dataset for CIFAR-10 compared to the other ansatzes. Across both datasets, our CNN-QNL-Net model significantly outperforms our PCA-QNL-Net model due to its efficient feature extraction capabilities before inputting the data into the QNL-Net architecture.

We note that our proposed PCA-QNL-Net models required higher  learning rates compared to our CNN-QNL-Net models. The proposed PCA-QNL-Net exhibited faster training in contrast to our CNN-QNL-Net, as CNNs entail an overhead for training parameters, utilizing a total of $34,282$ classical parameters for MNIST and $41,314$ classical parameters for CIFAR-10. Further, there are $5D+4r$ classical parameters in the quantum gates and $22$ classical parameters for the linear layers. Details of the gradient optimization calculations are provided in the Appendix.

Table~\ref{table:results_comparison} compares the outcomes from our models to the benchmark quantum binary classifiers: QTN-VQC~\cite{qtnvqc2023}, Hybrid TTN-MERA~\cite{Grant2018-dy}, Tensor Ring VQC~\cite{Peddireddy2022TensorRP}, SQNN~\cite{wu2022scalable}, and QF-hNet-BN~\cite{Jiang2021-of}, where QTN-VQC,  Hybrid TTN-MERA, and Tensor Ring VQC have results in their papers on classes 0 and 1 in MNIST; and SQNN and QF-hNet-BN have results for classes 3 and 6 in their papers. Our hybrid classical-quantum models, particularly the proposed CNN-QNL-Net, outperforms the benchmark models across both datasets. This superior performance underscores the effectiveness of integrating classical convolutional networks with quantum neural network layers, enabling more robust and accurate classification.
\begin{table*}
\small 
\centering
\caption{Performance of the proposed QNL-Net model  on binary classification tasks across datasets: MNIST digits $0$  and $1$ and CIFAR-10 classes $2$ (bird) and $8$ (ship).}\label{table:results}
\begin{tabular}{|c|c|c|c|c|c|}
\hline
\bf{Dataset} & \bf{Ansatz} & \bf{Model} & \bf{Learning Rate} & \bf{Average Train Accuracy (\%)} & \bf{Average Test Accuracy (\%)} \\
\hline
& 0 & CNN-QNL-Net & $1 \times 10^{-4}$ & $99.97 \pm 0.02$& $99.96 \pm 0.03$\\
& 1 & CNN-QNL-Net & $1 \times 10^{-4}$ & $99.96 \pm 0.02$& $99.95 \pm 0.02$\\
MNIST & 2 & CNN-QNL-Net & $1 \times 10^{-4}$ & $99.96 \pm 0.03$& $99.95 \pm 0.04$\\
(0, 1) & 0 & PCA-QNL-Net & $1.5 \times 10^{-4}$ & $99.65 \pm 0.17$& $99.54 \pm 0.16$\\
& 1 & PCA-QNL-Net & $1.5 \times 10^{-4}$ & $99.24 \pm 0.19$ & $99.18 \pm 0.34$\\
& 2 & PCA-QNL-Net & $1.5 \times 10^{-4}$ & $99.67 \pm 0.23$ & $99.59 \pm 0.21$\\
\hline
& 0 & CNN-QNL-Net & $3 \times 10^{-4}$ & $94.20 \pm 0.77$ & $93.54 \pm 0.66$\\
& 1 & CNN-QNL-Net & $3 \times 10^{-4}$ & $94.13 \pm 0.45$ & $93.98 \pm 0.37$\\
CIFAR-10 & 2 & CNN-QNL-Net & $3 \times 10^{-4}$ & $94.21 \pm 0.32$& $93.76 \pm 0.14$\\
(2, 8) & 0 & PCA-QNL-Net & $4 \times 10^{-4}$ & $81.94 \pm 1.51$& $81.16 \pm 1.09$\\
& 1 & PCA-QNL-Net & $4 \times 10^{-4}$ & $81.79 \pm 0.34$ & $80.95 \pm 0.35$\\
& 2 & PCA-QNL-Net & $4 \times 10^{-4}$ & $81.67 \pm 0.73$ & $80.86 \pm 0.74$\\
\hline
\end{tabular}
\vspace{-.1in}
\end{table*}
\begin{table}
\small
\centering
\caption{Performance of the QNL-Net model compared with QTN-VQC~\cite{qtnvqc2023}, Hybrid TTN-MERA~\cite{Grant2018-dy}, Tensor Ring VQC~\cite{Peddireddy2022TensorRP}, SQNN~\cite{wu2022scalable}, and QF-hNet-BN~\cite{Jiang2021-of} on binary classification tasks using the MNIST dataset. The CNN-QNL-Net model demonstrates better results using significantly fewer qubits.}\label{table:results_comparison}
\begin{tabular}{|l|c|c|c|}
\hline
\bf{Model} & \bf{Classes} & \bf{Qubits} & \bf{Test Accuracy}\\
\hline
QTN-VQC & $0, 1$ & 12 & $98.60$ \\
Hybrid TTN-MERA & $0, 1$ & 8 & $99.87 \pm 0.02$ \\
Tensor Ring VQC & $0, 1$ & 4 & $99.30$ \\
\bf{CNN-QNL-Net [ours]} & $0, 1$ & \bf{4} & $\bf{99.96 \pm 0.03}$\\
\hline
SQNN & $3, 6$ & $64$ & $97.47$\\
QF-hNet-BN & $3, 6$ & $12$ & $98.27$\\
\bf{CNN-QNL-Net [Ours]} & $3, 6$ & \bf{4} & $\bf{99.94 \pm 0.02}$\\
\hline
\end{tabular}
\vspace{-.2in}
\end{table}

For the MNIST dataset, the results depicted in Figure~\ref{fig:results_plots} indicate that the proposed PCA-QNL-Net model achieves a lower final loss compared to our proposed CNN-QNL-Net model across all ansatz configurations. This suggests that the PCA-QNL-Net model closely fits the training data by the end of the training period. However, the CNN-QNL-Net model attains a higher test accuracy more rapidly than the PCA-QNL-Net model, as illustrated in Figure~\ref{fig:results_plots}, indicating a superior generalization. In contrast, with respect to the CIFAR-10 data set, the CNN-QNL-Net model exhibits a faster reduction in training loss and achieves a lower final loss value compared to the PCA-QNL-Net model (Fig. \ref{fig:results_plots}b). This implies faster convergence and better fit to the CIFAR-10 training data. Furthermore, the CNN-QNL-Net model consistently achieves higher test accuracy with more speed than our PCA-QNL-Net model in Fig.~\ref{fig:results_plots}, reinforcing its efficacy in learning from complex data.

Overall, our experimental findings validate that the integration of quantum circuits with classical neural network architectures significantly enhances the performance of the QNL-Net model, establishing a new benchmark in quantum machine learning.


\section{Discussion} 
\label{sec:discuss}

The proposed QNL-Net leverages quantum entanglement as a key advantage. Although classical non-local blocks typically involve matrix multiplications within the embeddings and element-wise summation with the raw data, the QNL-Net utilizes $C_X$ entanglements to replicate variable dependencies present in classical non-local mechanisms. This fosters a highly interconnected system among all qubits, effectively extracting intricate probabilities and exploring complex data structures within a quantum framework. The specific pattern for entanglement in different ansatzes is not as crucial as the entanglement of all qubits with different rotations, creating a highly entangled state. Furthermore, applying another $R_z$ gate to the qubit, $0$, introduces phase modulation within the entangled system, crucial for fine-tuning the quantum state prior to measurement, thus influencing the probability of an outcome.

Despite the promising results obtained from these models, several limitations were identified. The current implementation is restricted by the reliance on classical computing methods that might be computationally inefficient as the datasets become much more extensive and the models are more complex. Still, it is a trade-off we must consider as we utilize lesser quantum resources. Multi-class classification posed a particular challenge as it performed poorly, likely due to the small circuit size with Pauli-$Z$ measurement at only one qubit, which limits the model's capability in distinguishing between multiple classes. Furthermore, the performance of the model could be further validated by testing on larger and more diverse datasets and exploring the effects of different quantum encodings and variational circuit designs.

Nevertheless, the implications of this research extend to various practical applications. In fields like image classification, medical imaging, and real-time video analysis, efficiently capturing long-range dependencies using quantum-enhanced models can lead to significant advancements in accuracy and performance. Furthermore, the scalable nature of our QNL-Net suggests that, as quantum hardware evolves, these models could be deployed in real-world scenarios, providing a competitive edge over traditional classical approaches.

\section{Conclusion} 
\label{sec:conclud}

This work introduces a Quantum Non-local Neural Networks (QNL-Net) framework as a novel hybrid classical-quantum approach for image classification. Through experiments on MNIST and CIFAR-10 datasets, our proposed QNL-Net models demonstrate competitive performance in binary classification tasks, utilizing fewer qubits compared to traditional quantum classifiers. The incorporation of fundamental quantum entanglement and rotation gates proves to be effective in capturing intricate spatial dependencies crucial for image analysis. 

\bibliography{references}

\newpage
\clearpage

\section{Appendix}

\subsection{Introduction to Quantum Computing}

Quantum computing utilizes the principles of quantum mechanics, such as superposition and entanglement, to enhance traditional computing capabilities. At the core of quantum computing are qubits, the quantum counterparts to classical bits. These qubits can simultaneously exist in states of $\ket{0}$ and $\ket{1}$ due to superposition, providing a significant increase in computational power compared to classical bits~\cite{nisqpreskill2018}, can be represented as 
\begin{eqnarray}
    \ket{0} = \begin{bmatrix} 1 \\ 0 \end{bmatrix}, \ket{1} =  \begin{bmatrix} 0 \\ 1  \end{bmatrix}. 
\end{eqnarray}

A qubit embodies a linear combination of the computational basis states, illustrating the fundamental concept of superposition. This state can be succinctly depicted as a vector within a two-dimensional complex Hilbert space, characterized by the following mathematical expression:
\begin{eqnarray} \label{eqn:qubitvector}
    \ket{\psi} = \alpha \ket{0} + \beta \ket{1},
\end{eqnarray}
where $\alpha, \beta \in \mathbb{C}$ are the complex coefficients of the quantum states $\ket{0}$ and $\ket{1}$ respectively. The probabilities of the qubit being in state $\ket{0}$ or $\ket{1}$ are given by the magnitude squared of these coefficients $|\alpha|^2$ and $|\beta|^2$. These probability amplitudes satisfy the normalization condition $|\alpha|^2 + |\beta|^2 = 1$.

Entanglement is another quantum phenomenon where the states of two or more qubits become interconnected, and the state of one qubit affects the other entangled qubits. This also demonstrates that the states cannot be factored into a product of individual qubit states as they are strongly correlated (i.e., $\ket{\psi_{AB}} \neq \ket{\psi_A} \otimes \ket{\psi_B}$ for states $A$ and $B$).


Quantum computations are performed primarily by manipulating quantum states through unitary transformations, achieved using quantum gates. Hadamard (H) gate is used to attain an equal superposition of the two basis states. The H gate maps the basis state $\ket{0}$ to $\frac{\ket{0} + \ket{1}}{\sqrt{2}}$ and the basis state $\ket{1}$ to $\frac{\ket{0} - \ket{1}}{\sqrt{2}}$. Rotation gates ($R_x, R_y, R_z$) rotate the state of a qubit around a specified axis on the Bloch sphere. A Phase ($P$) gate shifts the phase of a qubit by a specified angle $\lambda$, such that, applying $P(\lambda)$ to $\ket{\psi}$ in eq.(\ref{eqn:qubitvector}) results in 
\begin{eqnarray}
    P(\lambda)\ket{\psi} = \alpha \ket{0} + \beta e^{i \lambda} \ket{1}.
\end{eqnarray}
A CNOT ($C_X$ gate) is a two-qubit gate that flips the state of the second qubit (target) only if the first qubit (control) is $\ket{1}$. The following are the matrix representations of the relevant gates utilized in this work:
\begin{eqnarray}
    H = \frac{1}{\sqrt{2}} \begin{bmatrix} 1 & 1 \\ 1 & -1 \end{bmatrix}, \hspace{2mm} P(\lambda) = \begin{bmatrix} 1 & 0 \\ 0 & e^{i \lambda} \end{bmatrix},
\end{eqnarray}
\begin{eqnarray}
    R_x(\lambda) = \begin{bmatrix}
        cos(\frac{\lambda}{2}) & -i\hspace{1mm}sin(\frac{\lambda}{2}) \\
        -i\hspace{1mm}sin(\frac{\lambda}{2}) & cos(\frac{\lambda}{2})
    \end{bmatrix},
\end{eqnarray}
\begin{eqnarray}
    R_y(\lambda) = \begin{bmatrix}
        cos(\frac{\lambda}{2}) & -sin(\frac{\lambda}{2}) \\
        sin(\frac{\lambda}{2}) & cos(\frac{\lambda}{2})
    \end{bmatrix}, 
\end{eqnarray}
\begin{eqnarray}
    R_z(\lambda) = \begin{bmatrix}
        e^{-i \frac{\lambda}{2}} & 0 \\
        0 & e^{i \frac{\lambda}{2}}
    \end{bmatrix}, \hspace{2mm}
    C_X = \begin{bmatrix} 1 & 0 & 0 & 0 \\ 0 & 1 & 0 & 0 \\ 0 & 0 & 0 & 1 \\ 0 & 0 & 1 & 0 \end{bmatrix}.
\end{eqnarray}

\subsection{CNN Architecture}
We use the Convolutional Neural Network (CNN) architecture in combination with our QNL-Net Module because CNN is adept at capturing spatial dependencies and identifying local patterns within complex image data through convolutional and pooling layers, which reduce the input dimensionality while retaining essential features \cite{oshea2015introduction}.

In the proposed CNN-QNL-Net architecture, we start with two convolutional layers, each with an activation function and max pooling, for an input image tensor $X \in \mathbb{R}^{W \times H \times C}$, where $W$ is the width, $H$ is the height, and $C$ is the number of channels (i.e., 1 for grayscale images and 3 for RGB images) of the input image. In general, mathematically, a convolution operation `$*$' for an input image $I$ and a filter $K$ to output a feature map $F$ looks like,
\begin{eqnarray}
    F[i, j] = (I * K)_{[i, j]};
\end{eqnarray}
\begin{eqnarray}
    F[i, j] = \sum^{M-1}_{m=0} \sum^{N-1}_{n=0} \sum^{C-1}_{c=0} I_{[i+m,j+n,c]} \cdot K_{[m,n,c]}, 
\end{eqnarray}
where $i, j$ are positions in the output feature map $F$, and $m, n$ are positions in the filter $K$ for channel $c$. $M$, $N$, and $C$ are the width, height, and number of channels of the filter, respectively. The first convolutional layer applies a 2D convolution operation with $K_1$ filters (or kernels) of size $5 \times 5$ resulting in $K_1$ output channels, and is defined as,
\begin{eqnarray}
    Y_{[k]} = \sum^{C}_{c=1} X_{[c]} * W_{[k]} + b_{[k]}, \hspace{3mm} k = 1, ..., K_1,
\end{eqnarray}
where $W_{[k]}$ is the $k$-th filter weight and $b_{[k]}$ is the bias term. We apply the activation function $ReLU$, which simply eliminates the negative values in an input vector and is defined as $ReLU(x) = max(0, x)$, on each filter element-wise such that,
\begin{eqnarray}
    A_{[k]} = ReLU(Y_{[k]}), \hspace{3mm} A \in \mathbb{R}^{W_1 \times H_1 \times K_1},
\end{eqnarray}
where $W_1$ and $H_1$ are the width and height after convolution. Then, we apply a max pooling operation with a pool size of $2 \times 2$ on the Convolution + ReLU layer to obtain the pooled tensor $P$, which reduces the spatial dimensions of each channel by selecting the maximum value within each pool, such that,
\begin{eqnarray}
    P_{[k]} = MaxPool(A_{[k]}), \hspace{3mm} P \in \mathbb{R}^{W_2 \times H_2 \times K_1},
\end{eqnarray}
where $W_2 = \frac{W_1}{2}$ and $H_2 = \frac{H_1}{2}$. This Convolution + ReLU + MaxPool layer combination is repeated again on $P$ for further feature reduction with $K_2$ filters of size $5 \times 5$, and we obtain,
\begin{eqnarray}
    Z_{[k]} = \sum^{K_1}_{c=1} P_{[c]} * W_{[k]} + b_{[k]}, \hspace{3mm} k = 1, ..., K_2;
\end{eqnarray}
\begin{eqnarray}
    B_{[k]} = ReLU(Z_{[k]}), \hspace{3mm} B \in \mathbb{R}^{W_3 \times H_3 \times K_2},
\end{eqnarray}
where $W_3$ and $H_3$ are the width and height after the second convolution. Again, we apply the max pooling operation with a pool size of $2 \times 2$ on the previous layer,
\begin{eqnarray}
    Q_{[k]} = MaxPool(B_{[k]}), \hspace{3mm} Q \in \mathbb{R}^{W_4 \times H_4 \times K_2},
\end{eqnarray}
where $W_4 = \frac{W_3}{2}$ and $H_4 = \frac{H_1}{2}$. We apply a Dropout layer \cite{JMLR:v15:srivastava14a} to the resultant pooled tensor $Q$ to prevent overfitting which sets any element of the input to 0 during training with probability $p$, such that,
\begin{eqnarray}
    D = Dropout(Q, p), \hspace{3mm} p = 0.5.
\end{eqnarray}
Further, we flatten the output $D$ to obtain a 1-dimensional vector $F$ of size $(W_4 \cdot H_4 \cdot K_2) \times 1$. We apply a fully-connected (FC) layer to this flattened vector $F$ with ReLU activation, which results in,
\begin{eqnarray}
    H_1 = ReLU(W_{1} F + b_1), \hspace{3mm} H_1 \in \mathbb{R}^{128 \times 1},
\end{eqnarray}
where $W_1$ is the weight matrix and $b_1$ is the bias vector. FC layers simply apply a linear transformation to an input vector, essential for dimensionality reduction, aggregating scattered patterns across the features, and optimizing parameters \cite{kocsis2022lowdataregime}. Further, applying a non-linear activation function (e.g., ReLU) to a linear transformation enables representing non-linear relationships within the data. Finally, to obtain an output vector $H_2$ with size $4 \times 1$, which can be passed to the QNL-Net module, we apply our last FC layer to the output of $H_1$ such that,
\begin{eqnarray}
    H_2 = W_{2} H_1 + b_2, \hspace{3mm} H_2 \in \mathbb{R}^{4 \times 1},
\end{eqnarray}
where $W_2 $ is the weight matrix and $b_2$ is the bias vector.

\subsection{Principal Component Analysis (PCA)}
Principal Component Analysis is another linear dimensionality reduction technique that is suitable for linearly separable datasets used in this study. It uses Singular Value Decomposition (SVD) of the data to project it to a lower dimensional space. PCA proves to be computationally efficient and easy to compute compared to a technique like CNN. It does provide some disadvantages by losing some patterns and information in the data when reducing its dimensionality \cite{Jolliffe2016-ax}.
In the PCA-QNL-Net architecture, our input data matrix is $X \in \mathbb{R}^{N \times P}$, where $N$ is the number of samples in the dataset and $P$ is the total number of pixels per image. Before applying the SVD, the input data is centered for each feature, such that,
\begin{eqnarray}
    \mu = \frac{1}{N} \sum_{i=1}^{N} X_i,
\end{eqnarray}
\begin{eqnarray}
    \bar{X} = X - 1_N \mu^T,
\end{eqnarray}
where $\mu$ is the calculated mean vector of the data, $1_N$ is an N-dimensional vector of ones, and $\bar{X}$ is the centered input data matrix. We perform SVD on this centered matrix to decompose it into several component matrices,
\begin{eqnarray}
    \bar{X} = U \Sigma W^T,
\end{eqnarray}
where $U \in \mathbb{R}^{N \times N}$ is a matrix with each of its columns being a length-$N$ orthogonal unit vector or the left singular vector of $X$, $\Sigma \in \mathbb{R}^{N \times P}$ is a diagonal matrix composed of singular values of $X$, and $W \in \mathbb{R}^{P \times P}$ is a matrix with each of its columns being a length-$P$ orthogonal unit vector or the right singular vector of $X$. We project the centered data matrix onto the principal components by selecting the desired $L$ number of columns (or principal components, i.e., 4 in this case) of $W$, such that,
\begin{eqnarray}
    Z = \bar{X} W_L,
\end{eqnarray}
where $Z \in \mathbb{R}^{N \times L}$ is the desired reduced form of the data. This matrix is standardized to have zero mean and unit variance,
\begin{eqnarray}
    \bar{Z} = \frac{Z - \mu_z}{\sigma_z},
\end{eqnarray}
where $\mu_z$ and $\sigma_z$ are the mean and standard deviation of $Z$ respectively. $\bar{Z}$ is a reduced vector of size $L \times 1$ (i.e., $4 \times 1$) and is then passed to a fully connected layer, such that,
\begin{eqnarray}
    H_3 = W_{3} \bar{Z} + b_3,
\end{eqnarray}
where $W_3$ is the weight matrix and $b_3$ is the bias vector. Finally, $H_3 \in \mathbb{R}^{4 \times 1}$ can be fed directly to the QNL-Net for further processing.

\subsection{Gradient Calculations}
In the proposed QNL-Net framework, a hybrid gradient backpropagation approach is used to train our model effectively. This approach comprises optimizing both the classical parameters present in the neural nets and the set of trainable classical parameters which are the angles of quantum gates in the VQC. This hybrid training approach first applies a forward pass to optimize parameters for the convergence of the loss function. Our model uses the negative log-likelihood (NLL) loss function for the binary classification problem. The NLL loss measures the variation between the true labels $y$ and the classical predicted probabilities $\hat{y} = [\hat{y}_0, \hat{y}_1]$ obtained from the measurement of the hybrid quantum-classical QNL-Net model, and is defined for binary classification as:
\begin{eqnarray}
    \mathcal{L}(\phi) = 
    -  \sum^{N}_{i=1} \left(y_i \log \hat{y}_{i1} + (1 - y_i) \log \hat{y}_{i0}\right),
\end{eqnarray}
where $N$ is the number of samples in the dataset, $\hat{y}_0$ is the predicted probability for class 0, $\hat{y}_1$ is the predicted probability for class 1, and $\phi$ are the classical parameters.

To optimize these parameters, we compute the gradients of the loss with respect to the predicted probabilities,
\begin{eqnarray} \label{eqn:41}
   \frac{\partial \mathcal{L}}{\partial \hat{y}_{i1}} = - \frac{y_{i}}{\hat{y}_{i1}}, \hspace{2mm} \frac{\partial \mathcal{L}}{\partial \hat{y}_{i0}} = - \frac{1 - y_{i}}{\hat{y}_{i0}}.
\end{eqnarray}

The gradients of the predicted probabilities are then computed using standard back-propagation techniques as follows,
\begin{eqnarray} \label{eqn:42}
    \frac{\partial \mathcal{L}}{\partial \phi} = 
    \sum_i \left(
    \frac{\partial \mathcal{L}}{\partial \hat{y}_{i1}} 
    \frac{\partial \hat{y}_{i1}}{\partial h_i} 
    + 
    \frac{\partial \mathcal{L}}{\partial \hat{y}_{i0}} 
    \frac{\partial \hat{y}_{i0}}{\partial h_i}
    \right) \frac{\partial h_i}{\partial \phi},
\end{eqnarray}
where $h$ denotes the output from the classical model.

These derived first-order objective functions are optimized using the stochastic gradient descent method, Adam \cite{kingma2017adam}. The first-order moment $m_t$ and the second-order moment $v_t$ for the gradient (from Eq. (\ref{eqn:42})) are estimated using the following equations,
\begin{eqnarray}
    m_t = \beta_1 m_{t-1} + (1 - \beta_1) \frac{\partial \mathcal{L}}{\partial \phi},
\end{eqnarray}
\begin{eqnarray}
    v_t = \beta_2 v_{t-1} + (1 - \beta_2) \left(\frac{\partial \mathcal{L}}{\partial \phi}\right)^2,
\end{eqnarray}
where $t$ is the iteration/time-step and constants $\beta_1$ \& $\beta_2$ are the exponential decay rate. These moments are corrected for initialization bias, and we obtain bias-corrected moments such that,
\begin{eqnarray}
    \hat{m}_t = \frac{m_t}{1 - \beta_1^t}, \hspace{2mm} \hat{v}_t = \frac{v_t}{1 - \beta_2^t}.
\end{eqnarray}
Then, the parameters are updated accordingly,
\begin{eqnarray}
    \phi \leftarrow \phi - \eta \frac{\hat{m}_t}{\sqrt{\hat{v}_t} + \epsilon}, 
\end{eqnarray}
where $\eta$ is the learning rate and $\epsilon$ is an added small constant for numerical stability.

We also utilize the ExponentialLR scheduler \cite{li2019exponential} to adjust the learning rate $\eta$ after every epoch $t$ for faster convergence to obtain,
\begin{eqnarray}
    \eta_{new} = \eta \cdot \gamma^t,
\end{eqnarray}
where $\eta_{new}$ is the updated learning rate and $\gamma$ is the decay rate.

\subsection{Datasets Utilized}

MNIST \cite{lecun1998} is a handwritten digit recognition dataset used for many machine learning and computer vision tasks. Each image in MNIST is a grayscale 28 x 28-pixel representation of handwritten digits ranging from 0 to 9. The MNIST dataset contains 60,000 training samples used to train models and 10,000 testing samples used to evaluate model performance. These samples are handwritten by various individuals, covering a lot of variations and styles, ideal for machine learning. 

CIFAR-10 \cite{alexnet2012} is another widely-used benchmark dataset in the field of computer vision. It presents a collection of 32 x 32 size RGB images distributed across ten classes, including images of objects such as airplanes, cars, birds, cats, etc. The dataset contains a total of 50,000 training samples (5000 training samples per class) and 10,000 testing samples (1000 testing samples per class). Its diverse set of classes, coupled with variations in lighting, angle, and pose within images, makes it a suitable dataset for evaluating the robustness and generalization capability of image classification models.

\begin{figure}[h]
    \centering
    \includegraphics[width=0.3\textwidth]{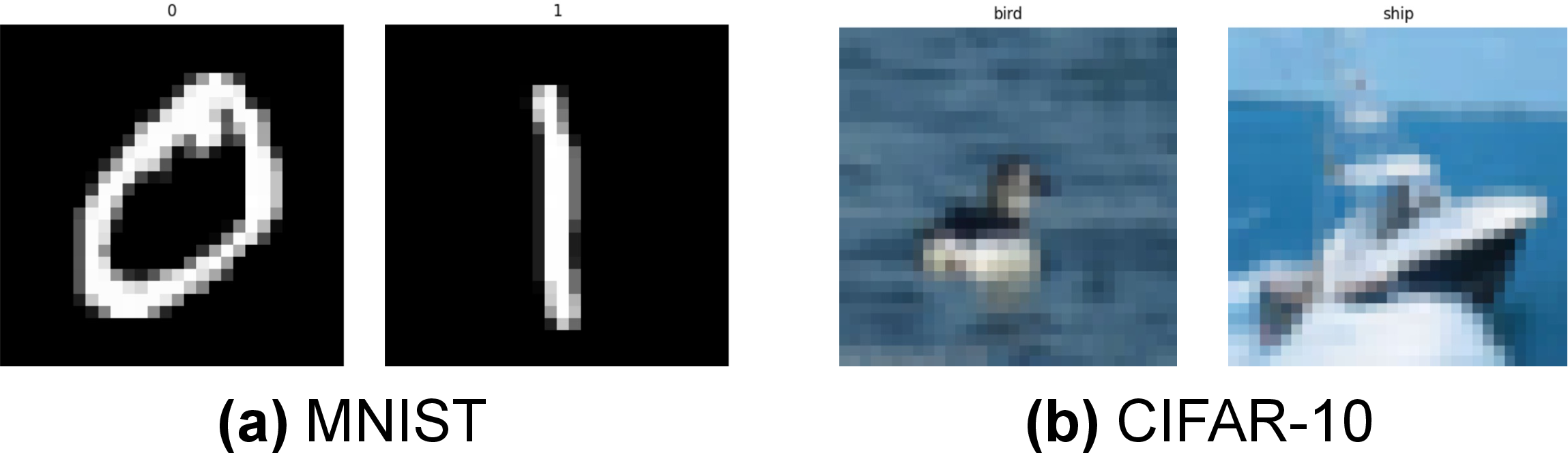}
    \caption{Sample images from the datasets used: MNIST \cite{lecun1998} (classes 0 and 1) and CIFAR-10 \cite{alexnet2012} (classes 2 and 8, i.e., bird and ship respectively).}
    \label{fig:dataset_sample}
\end{figure}
\end{document}